\documentclass[review]{elsarticle}

\usepackage{hyperref}

\usepackage{graphicx}
\usepackage{subfigure}
\journal{Journal of \LaTeX\ Templates}
\usepackage{framed}
\usepackage{algorithm}
\usepackage{bm}
\usepackage{amsmath}
\usepackage{float}
\usepackage{multirow}
\usepackage{color}
\usepackage[noend]{algpseudocode}
\usepackage{booktabs}
\usepackage{threeparttable}
\usepackage{tcolorbox}
 
\usepackage{algorithmicx}
\usepackage{algpseudocode}
\usepackage{soul}
\begin{document}

\begin{frontmatter}

\title{GIU-GANs:Global Information Utilization for Generative Adversarial Networks }


\author[mymainaddress,thirdaddress]{Yongqi Tian}
\ead{3220190462@bit.edu.cn}

\author[mysecondaryaddress]{Xueyuan Gong\corref{mycorrespondingauthor}}
\cortext[mycorrespondingauthor]{Corresponding author}
\ead{xygong@jnu.edu.cn}

\author[thirdaddress,forthaddress]{Jialin Tang}
\ead{01068@bitzh.edu.cn}

\author[mymainaddress,thirdaddress]{Binghua Su}
\ead{bhsu@263.net}

\author[mysecondaryaddress]{Xiaoxiang Liu}
\ead{tlxx@jnu.edu.cn}

\author[mysecondaryaddress]{Xinyuan Zhang}
\ead{Zhangxy@jnu.edu.cn}

\address[mymainaddress]{School of Optoelectronics, Beijing Institute of Technology, Beijing, China}
\address[mysecondaryaddress]{School of Intelligent Systems Science and Engineering, Jinan University, Zhuhai, China}
\address[thirdaddress]{School of Information Technology, Beijing Institute of Technology, Zhuhai, China}
\address[forthaddress]{Faculty of Data Science, City University of Macau, Macau, China}

\begin{abstract}
Recently, with the rapid development of artificial intelligence, image generation based on deep learning has advanced significantly. Image generation based on Generative Adversarial Networks (GANs) is a promising study. However, because convolutions are limited by spatial-agnostic and channel-specific, features extracted by conventional GANs based on convolution are constrained. Therefore, GANs cannot capture in-depth details per image. Moreover, straightforwardly stacking of convolutions causes too many parameters and layers in GANs, yielding a high overfitting risk. To overcome the abovementioned limitations, in this study, we propose a GANs called GIU-GANs (where Global Information Utilization: GIU). GIU-GANs leverages a new module called the GIU module, which integrates the squeeze-and-excitation module and involution to focus on global information via the channel attention mechanism, enhancing the generated image quality. Moreover, Batch Normalization (BN) inevitably ignores the representation differences among noise sampled by the generator and thus degrades the generated image quality. Thus, we introduce the representative BN to the GANs’ architecture. The CIFAR-10 and CelebA datasets are employed to demonstrate the effectiveness of the proposed model. Numerous experiments indicate that the proposed model achieves state-of-the-art performance.
\end{abstract}

\begin{keyword}
\texttt{}Image Generation, Generative Adversarial Networks, Global Information Utilization, Involution, Representative Batch Normalization

\end{keyword}

\end{frontmatter}

\section{Introduction}

Image generation is crucial and challenging in the Computer Vision field. With the development of deep learning, there has been remarkable progress in this field. In 2014, the application of Generative Adversarial Networks (GANs) \cite{DBLP:journals/corr/GoodfellowPMXWOCB14} for image generation was reported with promising results. Then, GANs based on deep Convolutional Neural Networks (CNNs) \cite{DBLP:journals/corr/RadfordMC15} improved the generated image quality. Since then, CNNs have been the core of GANs. However, convolution kernels have two remarkable properties, spatial-agnostic and channel-specific \cite{DBLP:journals/corr/abs-2103-06255}, which contribute to their widespread yet pose some issues.

Spatial-agnostic means that a convolution kernel produces the same output no matter its location in an image. As is well known, the size of a convolution kernel is $C_o \times C_i\times K\times K$, where $C_i$ and $C_o$ are the numbers of input and output channels, respectively, and $K$ is the kernel size. Owing to spatial-agnostic, a convolution kernel has the same parameters in different regions of an image, which reduces computation. However, this constrains the receptive field of convolution, which makes it difficult for models to capture the mathematical relationships of spatially distant locations and thus deprives the ability of convolution to extract global information. The convolution kernel of each channel has specific parameters so that CNNs can detect different features, which is the second property of convolution called channel-specific. Generally, the number of channels increases as a network deepens. Yet, it is questionable if more channels yield higher accuracy. When the quality of a training set is poor when models are complex, there may be overfitting risks that lead to poor performance. Hundreds of channels complicate a model and increase the overfitting risks. It has been shown that many convolution kernels are redundant in the channel dimension \cite{DBLP:conf/bmvc/JaderbergVZ14}, so the traditional convolution operation may increase computation yet fail to improve the model performance.

For CNN-based GANs, spatial-agnostic has a low time complexity, yet it limits the Global Information Utilization (GIU) by GANs. Therefore, GANs often require stacking convolutions, resulting in too many layers. However, complex models increase the overfitting risk and reduce the generated image quality. Moreover, convolution gives the same response to different regions of an image, limiting the model adaptation to different regions of the image \cite{DBLP:journals/corr/abs-2103-06255}. This will limit the ability of CNN-based GANs to extract image details. To verify our view, we visualized the hidden layer of the discriminator of GANs using Principal Component Analysis. First, we subtract the mean of each dimensional feature of the input feature map $X$ and calculate its covariance matrix $\dfrac{1}{n}XX^T$, where $n$ denotes the number of samples. Second, we employ the eigenvalue decomposition method to find the eigenvalues and eigenvectors of the covariance matrix $\dfrac{1}{n}XX^T$. Then, we sort the eigenvalues in descending order and select the largest $K$ eigenvalues among them. To better observe the extraction of details, we choose $K=1$, i.e., to generate a gray-scale map to observe the results of its feature capture. Next, $K$ feature vectors are used as row vectors to form a feature vector matrix $P$. Finally, data are transformed into a new space constructed by the $K$ feature vectors. Figure 1 shows the hidden layers of discriminators for Least Squares GANs (LSGANs) \cite{DBLP:conf/iccv/MaoLXLWS17}, Wasserstein GAN with Gradient Penalty (WGAN-GP) \cite{DBLP:conf/nips/GulrajaniAADC17}, Spectrally Normalized GANs (SNGANs) \cite{DBLP:conf/iclr/MiyatoKKY18}, and GIU-GANs (our proposed model) on the CelebA dataset.

$L_D=-E_{x\sim p_{data(x)}}[logD(X)]-E_{z\sim p_z}[log(1-D(G(z)))]-E_{(x,y)\sim p_{data}(x,y)}[logp(y|x,y \textless K+1)]$
\begin{figure}[H]
	\centering
	\vspace{-5.0em}
	\includegraphics[width=14.85cm,height=9cm]{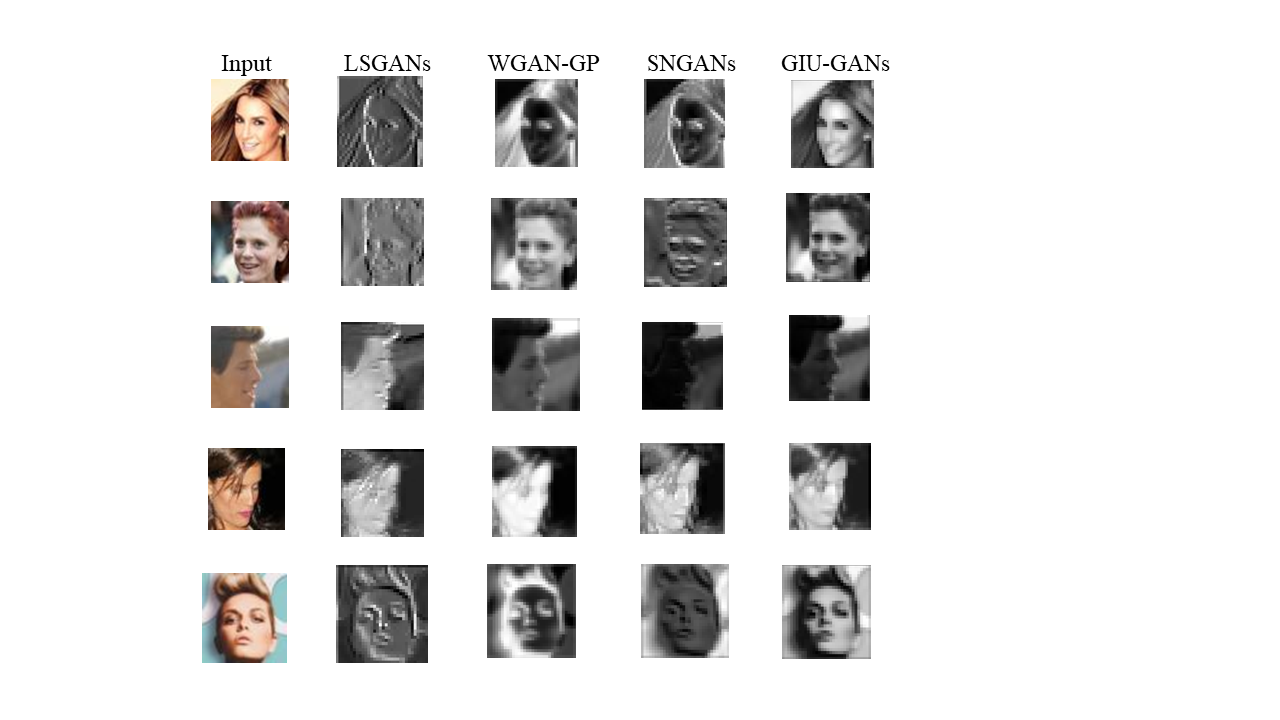}
	\vspace{-4.0em}
	\caption{Visualization of discriminator hidden layer for LSGANs, WGAN-GP, SNGANs, and GIU-GANs on CelebA.}
	
\end{figure}

Considering the channel-specific property, which causes redundant channels in a model. Redundant channels complicate GANs, resulting in an overfitting phenomenon. The overly complex discriminator makes a model have more stringent evaluation criteria for the generated false images. This will cause a severe problem for GANs, i.e., mode collapse, where the generator produces many similar images that lose diversity. In summary, how to appropriately improve the generated image quality without the overfitting risk is challenging.

Aiming to solve the problems of CNN-based GANs, we propose the GIU module for GANs to enhance the generated image quality. The GIU module comprises involution \cite{DBLP:journals/corr/abs-2103-06255} and the Squeeze-and-Extraction (SE) module \cite{DBLP:conf/cvpr/HuSS18}. Involution is a new neural network operator that contradicts convolution characteristics. It shares a kernel on the channel space and generates the kernel based on the input feature map. The size of the involution kernel is $H\times W\times K\times K\times G$, where $H\times W$ is the size of a single feature; $K$ is the kernel size, we set it to 3 in this study (equivalent to the conventional $3\times 3$ convolution); $G$ represents the number of groups. Thus, the feature map is divided into $G$ groups, with $\dfrac{C_i}{G}$ channels in each group ($G \ll C_i$), and the same involution kernel is shared within a group. Involution divides the channels of an input feature map into $G$ groups, with $\dfrac{C_i}{G}$ channels in a group, and extracts a set of points in the neighborhood $K\times K$ size of a certain pixel within the group for multiplication--addition operation with the generated involution kernel. Therefore, the point set of a pixel in a group and the involution kernel multiplication--addition does not change the number of output channels. For other feature extraction operations that do not change the number of output channels, such as $1\times 1$ convolution or $3\times 3$ convolution, the number of output channels needs to be the same as the number of input channels, i.e., $C_i=C_o$. However, the $G$ of involution is much smaller than the input channels. Using a smaller $G$ instead of $C_o$ can effectively avoid the generation of redundant channels and thus prevent overfitting. Meanwhile, the spatial-specificity of involution makes its kernel generated on a per-pixel basis on the feature map, thereby preventing it from producing the same response in a single feature map, i.e., all its kernels are not the same. Therefore, unlike conventional convolution, involution is more concerned with the diversity of different regions in a feature map, which is reflected in GANs by its ability to extract detailed features. Figure 1 confirms the superiority of GIU-GANs in terms of detailed features.

Moreover, the involution kernel’s parameter matrix is affected by the input feature channels. To generate a kernel parameter matrix that better fits the current task, GIU-GANs introduces the GIU module, which combined the channel attention mechanism SE module and involution together. The SE module uses two operations, i.e., squeeze and excitation, to explicitly model the interdependence between feature channels. The SE module uses the relationship between channels to obtain the importance degree of each channel through learning. It enhances the useful features according to this importance degree and prevents unuseful features for the current task. From the perspective of the number of parameters, the number of parameters in the GIU module is $\dfrac{(C_i^2+K^2 GC_i)}{r_1}+2C_i+\dfrac{(2C_i^2)}{r_2}$, where $\dfrac{(C_i^2+K^2 GC_i)}{r_1}$ is the number of involution parameters, $2C_i$ is the number of Batch Normalization (BN) parameters, $\dfrac{(2C_i^2)}{r_2}$ is the number of SE module parameters, $r_1$ is the ratio of channel reduction, which we set to 16, and $r_2$ is the channel scaling ratio for the SE module (we follow the original SE module \cite{DBLP:conf/cvpr/HuSS18} paper and set it to 16). Relative to the number of convolution parameters $k^2 C_i^2 C_o^2$, $C_o$ is the number of output channels, which is often set to hundreds or thousands of numbers, and we set $G$ much smaller than $C_o$. In summary, the number of parameters of the GIU module is much smaller than that of conventional convolution. We counted the number of parameters for all evaluated models and added them. According to the GIU module’s principle and parameter formula, it is suitable to be inserted after the “small feature map with more channels.” The experiment in Subsection 5.1 helps us verify this view.

For convolution, deeper networks can make the task of the convolutional kernel of each layer more explicit and thus effectively help the model fit the features better. For example, when the network is deeper, the first layer learns the edge features, the second layer learns the simple shapes, and the third layer learns the target’s complex shapes, whereas, when the network is shallower, the first layer learns the target’s complex shapes, which is difficult. Therefore, deeper models can have stronger feature extraction ability; motivated by this, we train several deeper WGAN-GP on the CelebA dataset and present the results in Figure 2 to observe the extraction of detailed features. $+2$ and $+3$ mean that two and three layers, respectively, have been added to both the WGAN-GP discriminator and generator. Figure 2 and Table 5 show that the WGAN-GP with three extra layers (parameters: 12.66M) achieved results comparable to those of the GIU-GANs (parameters: 6.91M). Through the GIU module, we could improve the feature extraction capability with fewer parameters for GANs and thus enhance the generated image quality.

\begin{figure}[H]
	\centering
	\includegraphics[width=14.85cm,height=9cm]{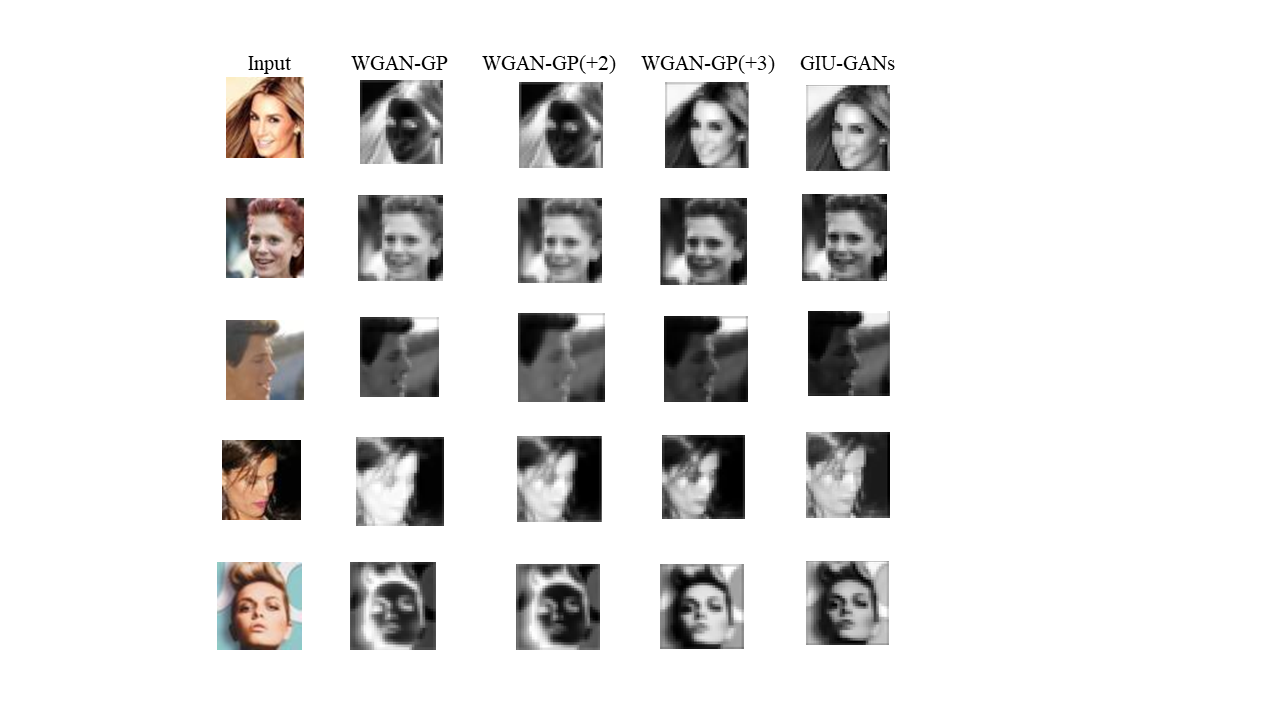}
	\vspace{-4.0em}
	\caption{Deeper visualization of hidden layers of WANG-GP and GIU-GANs on the CelebA dataset.}
\end{figure}

In GAN-based image generation tasks, random noises are sampled to generate images by the generator, and BN \cite{DBLP:conf/icml/IoffeS15} has become the core configuration of generators because of its stability training advantages. However, normalization causes generators to ignore the representation differences between different noises. To address this problem, we replace the BN in the generator with Representative Batch Normalization (RBN) \cite{DBLP:conf/cvpr/GaoHLCP21}. RBN introduces the formulation of centering and scaling calibrations to conventional BN. RBN compresses input features as representational features and enhances or suppresses them through learnable variables. Through RBN, GANs can enhance or suppress the representativeness difference of different random noises to enhance the generated image quality.

The contributions of this study can be summarized as follows.

\begin{itemize}
\item We design a new GIU module for GANs. The GIU module can exploit global information to improve the ability of GAN feature extraction and enhance the generated image quality.
\item We introduce RBN to GANs for more focus on the expression of representative features.
\item We adopt two technologies, spectral normalization \cite{DBLP:conf/iclr/MiyatoKKY18} and WGAN-GP \cite{DBLP:conf/nips/GulrajaniAADC17}, to stabilize the training of GANs and improve the generated image quality.
\end{itemize}

We compared our model with many classical GAN models, such as Deep Convolutional GANs (DCGANs) \cite{DBLP:journals/corr/RadfordMC15} LSGANs \cite{DBLP:conf/iccv/MaoLXLWS17}, WGAN-GP, SNGANs \cite{DBLP:conf/iclr/MiyatoKKY18}, and Self-Attention GANs (SAGANs) \cite{DBLP:conf/icml/ZhangGMO19} on CIFAR-10 and CelebA datasets. Each GAN was trained 1 million times, and all models generated 50,000 images every 100,000 times. The images were tested to calculate Inception Score (IS) \cite{DBLP:conf/nips/SalimansGZCRCC16} and Frechet Inception Distance (FID) \cite{DBLP:conf/iccv/LiuLWT15}. For all models, we selected the one with the best scores in 10 tests as the best model.

The rest of this article is organized as follows. Section 2 reviews studies related to GANs. In Section 3, we introduce GIU-GANs, including the architecture of the GIU module and techniques for stabilizing the training of GANs. The algorithm of RBN is presented in Section 4. In Section 5, we present the experimental results of many classical GAN models and our model implemented on the CIFAR-10 and CelebA datasets. Finally, Section 6 concludes this study.

\section{Related Work}

In this section, we review some previous studies related to GANs.

Image generation has long been a subject that has attracted researchers from a wide range of fields, including computer vision and machine learning. In 2014, GANs \cite{DBLP:journals/corr/GoodfellowPMXWOCB14} was proposed to generate images and has received extensive attention. Since then, various GANs have been developed. To enhance the generated image quality, Mirza et al. \cite{DBLP:journals/corr/MirzaO14} proposed conditional GANs, introducing the condition information as a label to the generator. In addition, Radford et al. \cite{DBLP:journals/corr/RadfordMC15} proposed DCGANs to improve GANs by full convolutional structure. Further, interpretable representation learning by information maximizing GANs \cite{DBLP:conf/nips/ChenCDHSSA16} constrained some dimensions of random noise to control the semantic characteristics of generated data. In addition, variational autoencoder with GAN \cite{DBLP:conf/icml/LarsenSLW16} introduced a discriminator basis on the conventional variational autoencoder \cite{DBLP:journals/corr/KingmaW13} to better capture the data distribution. In addition, Energy-based GAN \cite{DBLP:conf/iclr/ZhaoML17} introduced an autoencoder to the discriminator, showing an improved convergence pattern and scalability to generate high-resolution images. While the theory of GANs has made rapid progress, the applications of GANs have also flourished. The applications include text-to-image tasks \cite{DBLP:conf/icml/ReedAYLSL16}, \cite{DBLP:conf/iccv/ZhangXL17}, \cite{DBLP:journals/pami/ZhangXLZWHM19}, \cite{DBLP:conf/cvpr/HongYCL18}, image-to-image tasks \cite{DBLP:conf/icml/IoffeS15}, \cite{DBLP:conf/iccv/ZhuPIE17}, \cite{DBLP:conf/cvpr/KarrasLA19}, \cite{DBLP:conf/iclr/NieNP19}, and image super-resolution tasks \cite{DBLP:conf/cvpr/LedigTHCCAATTWS17}, \cite{DBLP:conf/eccv/WangYWGLDQL18}, \cite{DBLP:conf/eccv/BulatYT18}.

Despite GANs having made significant progress, the training of GANs is still filled with difficulties. Poorly trained GANs may generate similar images and reduce the diversity of generated images, leading to mode collapse. To address the issue, multiagent diverse GANs\cite{DBLP:conf/cvpr/GhoshKNTD18} built multiple generators simultaneously, and each generator produced a different pattern to guarantee the diversity of samples produced. Moreover, improving objective function is a strategy to stabilize the training of GANs. Unrolled GANs \cite{DBLP:conf/iclr/MetzPPS17} changed the objective function of the generator considering both the current state of the generator and discriminator after K updates. Meanwhile, Mao et al.\cite{DBLP:conf/iccv/MaoLXLWS17} proposed LSGANs to stabilize the training of GANs. LSGANs changed the objective function of GANs from cross-entropy loss to least square loss and achieved a certain effect. Further, Wasserstein GANs (WGANs) \cite{DBLP:conf/icml/ArjovskyCB17} replaced Jensen--Shannon (JS) divergence with Wasserstein distance, which considerably stabilized the training of GANs. In addition, because WGANs used the weight clipping strategy to enforce the Lipschitz constraint, the value of weight will affect the training result. Therefore, Gulrajani et al. \cite{DBLP:conf/nips/GulrajaniAADC17} proposed WGAN-GP, which replaced the strategy of weight clipping based on WGANs with the gradient penalty to meet the Lipschitz constraint for the training of GANs. Moreover, to satisfy the Lipschitz constraint, SNGANs \cite{DBLP:conf/iclr/MiyatoKKY18} used spectral normalization to stabilize the training of GANs. On this basis, SAGANs \cite{DBLP:conf/icml/ZhangGMO19} combined spectral normalization with self-attention \cite{DBLP:conf/nips/VaswaniSPUJGKP17} mechanism to improve the performance of GANs.

\section{Global Information Utilization for Generative Adversarial Networks (GIU-GANs)}

In this section, we present GIU-GANs in detail. Firstly, we introduce preliminary work on GIU-GANs, including SE Module and involution. Then, the implementation method of the GIU module will be explained. Next, we show the architecture of GIU-GANs. Finally, we describe two techniques in detail to stabilize the training of GANs, WGAN-GP and spectral normalization for both generator and discriminator.

\subsection{Preliminary Work}

In this subsection, we describe the channel attention SE module and involution in detail.

\subsubsection{Squeeze-and-Excitation Module (SE module)}

At present, CNNs are core components of Computer Vision research. Capturing more information from images is becoming increasingly relevant in CNNs. There has numerous studies \cite{DBLP:conf/cvpr/SzegedyVISW16}, \cite{DBLP:conf/eccv/NewellYD16}, considering the spatial domain to improve the model performance, whereas the SE module considers the relationship between characteristic channels. It captures the importance of each feature channel and then refers to this importance enhancement or suppression feature.

The first step is to compress each feature channel to a real number $z$ using global average pooling \cite{DBLP:journals/corr/LinCY13}. Each real number obtained by compression has a global field of sensation, and their number of channels matches the number of input channels.

Then, the SE module captures the relationship between the channels via the excitation operation. It employs a gating mechanism in the form of the sigmoid function:
\begin{equation}
	s=F_{ex}(z,w)=\sigma(g(z,W))=\sigma(W_2\delta(W_1z)),
\end{equation}
where $W_1\in R^{\frac{C}{r}\times C},W_2\in R^{C\times \frac{C}{r}}$. The SE module adopts full connection layer operation to reduce the dimension of $s$ by squeeze, which can reduce the computational load. Then, the result of dimensionality reduction is processed using the Rectified Linear Unit (ReLU) activation function, which is then multiplied by the full connection layer $W_2$. The output dimension is $1\times 1\times C$. Next, the feature maps are processed by the sigmoid function to obtain $s$.

Finally, the sigmoid activation value of each captured channel is multiplied by the original feature:
\begin{equation}
	\tilde x_c=F_{scale}(u_c,s_c)=s_c\cdot u_c.
\end{equation}

As such, useful feature channels are enhanced, and less useful feature channels are suppressed.

\subsubsection{Involution}

Involution, which is a new neural network operator, has different characteristics from convolution. Involution has the characteristics of spatial-specificity and channel-agnostic. The former property contributes to generating parameters of the kernel by feature maps to improve the ability of feature extraction. The latter property can reduce the number of channels, thereby significantly decreasing the computational load and preventing model overfitting. The involution kernel size is $H\times W\times K\times K\times G$, where $K$ is the size of the involution kernel $H$ and $W$ are resolutions of the feature map. Involution is defined as follows:  
\begin{equation}
	Y_{i,j,k}=\sum_{(u,v)\in \Delta k}H_{i,j,u+[K/2],v+[K/2],[kG/C]}X_{i+u,j+v,k},
\end{equation}
where $H\in R^{H\times W\times K\times K\times G}$ is the approach by which the involution kernel is created:
\begin{equation}
	H_{i,j}=\phi(X_{\Psi_{i,j}}).
\end{equation}
$\Psi_{i,j}$ is the index set in the neighborhood of pixel $(i,j)$ to obtain parameters of number $1\times 1\times C$, whereas $\phi$ is a series of scaling and reshape operations. Selecting a single point set of ${i,j}$ on the feature map can acquire involution kernel instantiation:
\begin{equation}
	H_{i,j}=\phi(X_{i,j})=W_1\sigma(W_0X_{i,j}),
\end{equation}
where $W_0\in R^{\frac{C}{r}\times C}$, $W_1\in R^{K\times K\times G\times \frac{C}{r}}$. Involution uses $W_0$ and $W_1$ to transform the parameter matrix to $K\times K\times G$; $r$ is the channel reduction ratio, and $\sigma$ is the BN and ReLU operations.

Involution turns the feature vector of pixel $(i, j)$ in a feature map into $K\times K\times G$ by $\phi$ and reshapes the operation, where $K\times K$ is the kernel shape, and $G$ is the number of shared kernels. Finally, involution performs multiplication--addition operations on the weight matrix of size $K\times K\times G$ and the feature vector of the neighborhood of pixel to obtain the final outputting feature map.

\subsection{Global Information Utilization (GIU) module}
In this subsection, we present the algorithm of the GIU module comprising the SE module and involution.
\subsubsection{Algorithm of GIU module}
Notably, the involution kernel generative function $\phi$ considers using the feature vector of point $(i,j)$ on the feature map for design:
\begin{equation}
	\phi(X_{i,j})=W_1\sigma (W_0X_{i,j}),
\end{equation} 
where $X_{i,j}$ is the feature vector of the pixel located in point $(i,j)$. The value of the feature vector represents the output of point $(i,j)$ on each channel. Different channel responses have different effects on the current task; useful values will improve the performance of GANs, whereas less useful values will misguide GANs. To obtain a more effective involution kernel, we combine the SE module with involution, called the GIU module, to adaptively enhance kernel useful parameters and suppress useless parameters.  Figure 3 is a schematic of the architecture of the GIU module.
\begin{figure}[H]
	\centering
	\includegraphics[width=9.9cm,height=6cm]{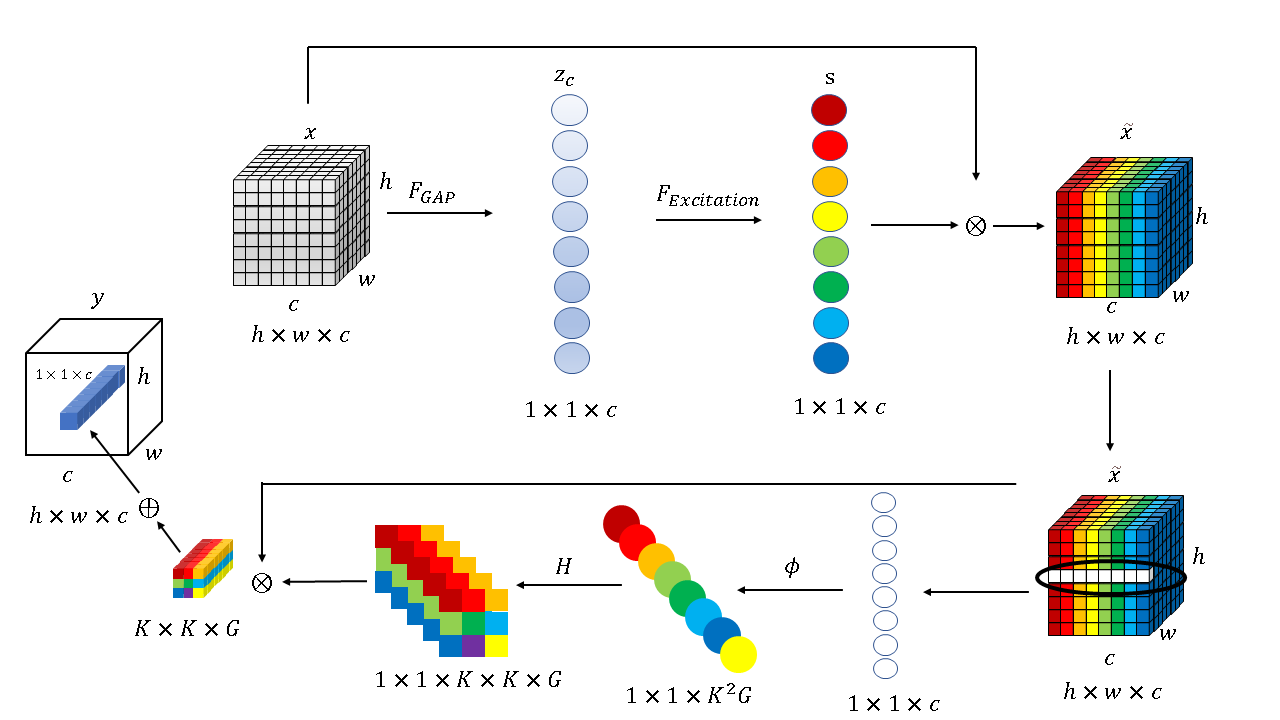}
	\vspace{-1.0em}
	\caption{A schematic of the GIU module for GIU-GANs. The $\otimes$ and $\oplus$ represent matrix multiplication and addition, respectively.}
\end{figure}
First, the SE module is employed to capture the importance of each channel for the input feature map:
\begin{equation}
	s=\sigma(W_2\delta(W_1F_{sq}(X))),
\end{equation} 
where $W_1\in R^{\frac{C}{r}\times C},W_2\in R^{C\times \frac{C}{r}}$. $F_{GAP}(X)$ is global average pooling operation: 
\begin{equation}
	z_c=F_{GAP}(X)=\frac{1}{H\times W}\sum_{i=1}^H\sum_{i=1}^Wu_c(i,j),
\end{equation}
$u_c$ are feature maps where input $X$ uses a convolution operation, $H$ and $W$ are spatial dimensions of feature maps, and $s$ is the importance of each channel. Multiplying the $s$ of each channel by itself, we have the following: $\tilde X=s\cdot u_c,$, where $u\in R^{H\times W\times C}$ is a two-dimensional spatial kernel and $u_c$ represents the c-th channel.

Then, the obtained $\tilde X$ is used to generate the involution kernel:
\begin{equation}
	\tilde H_{i,j}=\phi(\tilde X_{\Psi_{i,j}}).
\end{equation} 
The previous involution kernel is the point set at pixel $(i, j)$: $H_{i,j}=[x_c^1,x_c^2,...,x_c^{K^2}]$, and the improved kernel function is $\tilde H_{i,j}=[\tilde x_c^1,\tilde x_c^2,...,\tilde x_c^{K^2}]$. The new generation function can select channels, thereby adaptively adjusting the parameter matrix according to the importance of the channel. Reshaping the involution kernel size to $K\times K\times G$, where $G$ is the number of groups with each group sharing the same involution kernel, then the final feature map can be obtained by multiplication--addition using the feature vector of the GIU module and the neighborhood of the corresponding point on the input feature map:
\begin{equation}
	\tilde Y_{i,j,k}=\sum_{(u,v)\in \Delta k}\tilde H_{i,j,u+[K/2],v+[K/2],[kG/C]}X_{i+u,j+v,k}.
\end{equation} 
Specifically, the algorithm of the GIU module is shown in Algorithm 1.
\begin{algorithm}[t]
	\caption{Algorithm of GIU module} %算法的名字
	\hspace*{0.02in} {\bf Input: }
	The set of feature maps for current batch, $X_c$, its spatial dimensions are $H$ × $W$.\\ 
	$W_0\in R^{\frac{C}{r}\times C},W_1\in R^{C\times \frac{C}{r}}$  are full connection layer operations.\\
	$W_2\in R^{\frac{C}{r}\times C}$, $W_3\in R^{K\times K\times G\times \frac{C}{r}}$ are convolution operations.\\
	$\sigma$ is sigmoid activation function, $\delta$ is ReLU activation function.\\ %算法的输入， \hspace*{0.02in}用来控制位置，同时利用 \\ 进行换行
	\hspace*{0.02in} {\bf Output:} 
	The set of feature maps for current batch, $\tilde Y_{i,j,k}$%.算法的结果输出
	\begin{algorithmic}[1]
		\State $z_c\gets \frac{1}{H\times W}\sum_{i=1}^H\sum_{i=1}^WX_c(i,j)$.
		\State $s\gets \sigma(W_1\delta(W_0z_c))$.
		\State $\tilde X_c \gets s\cdot u_c$.
		\State $\tilde H_{i,j}\gets W_3\delta (W_2\tilde X_c)$.
		\State Reshaping the size of involution kernel: 1 × 1 × $K^2G$$\stackrel{reshape}{\longrightarrow}$ $K\times K\times G$.
		\State $\tilde Y_{i,j,k}\gets \sum_{(u,v)\in \Delta k}\tilde H_{i,j,u+[K/2],v+[K/2],[kG/C]}X_{i+u,j+v,k}$.
		\State \Return $\tilde Y_{i,j,k}$.
	\end{algorithmic}
\end{algorithm}

\subsection{The architecture of GIU-GANs}
The architecture of GIU-GANs is shown in Figure 4 and Figure 5.

\begin{figure}[H]
	\centering
	\includegraphics[width=9.9cm,height=6cm]{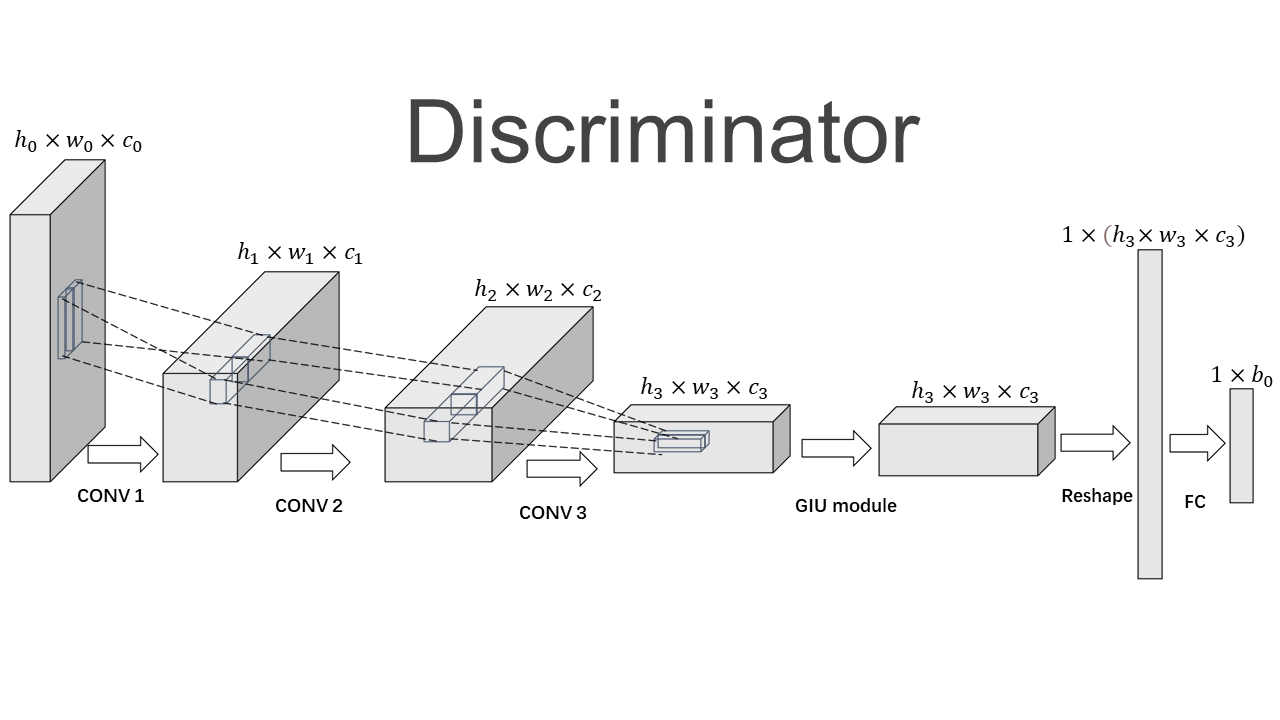}
	\vspace{-4.0em}
	\caption{The discriminator architecture of GIU-GANs. $h_x$ and $w_x$, respectively, denote the height and width of the current feature map, $c_x$ denotes the number of channels in the current hidden layer, and $b_0$ denotes the batch size.}
\end{figure}
\begin{figure}[H]
	\centering
	\includegraphics[width=9.9cm,height=6cm]{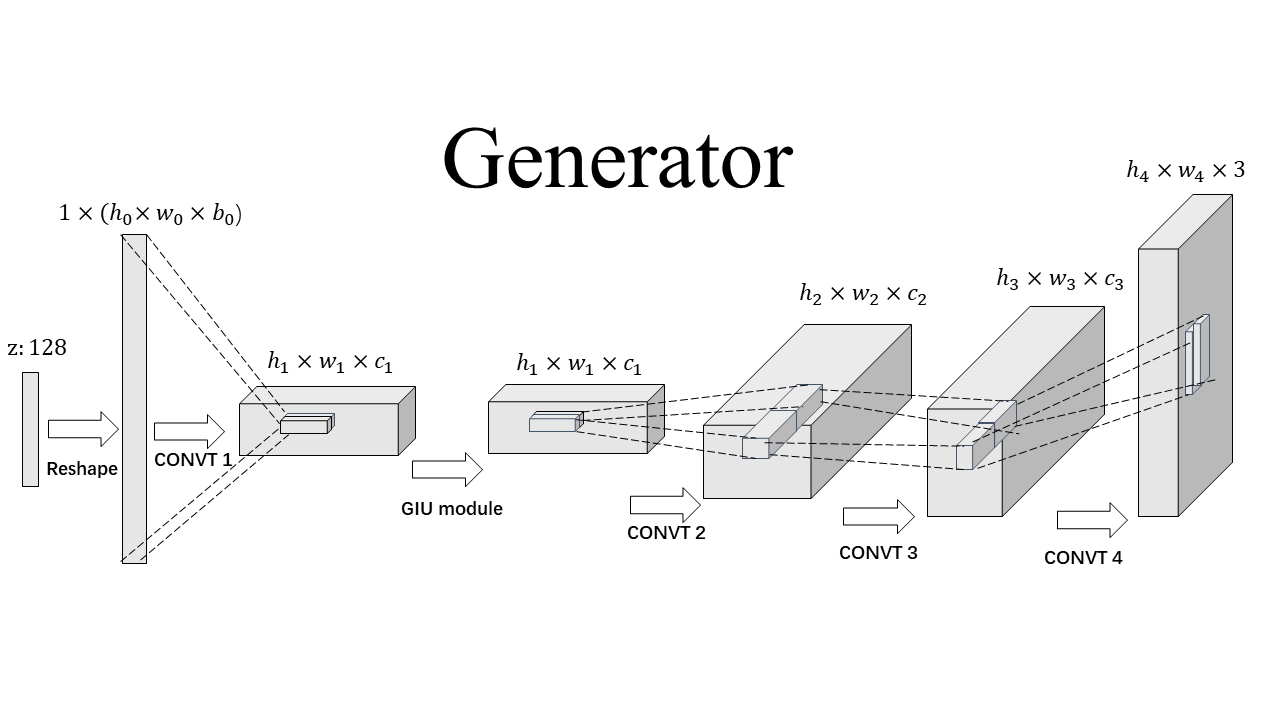}
	\vspace{-4.0em}
	\caption{The generator architecture of GIU-GANs. $h_x$ and $w_x$, respectively, denote the height and width of the current feature map, $c_x$ denotes the number of channels in the current hidden layer, and $b_0$ denotes the batch size.}
\end{figure}

The discriminator does not use BN, and GIU-GANs select LeakReLU with the slope of the leak set to 0.1 and ReLU for the discriminator and generator, respectively.

The GIU module focuses on the importance of the channels in the hidden layer and uses this as an important basis for generating the parameter matrix. Meanwhile, the formula of the number of parameters of the GIU module is expressed as follows:
\begin{equation}
	\frac{C^2_i+K^2GC_i+r_iC_i}{r_i}+\frac{2C^2_i}{r_2}.
\end{equation}
Considering the necessity of reducing the number of model parameters, we insert the GIU module into the latter layer of the input layer of the generator and the former layer of the output layer of the discriminator. The experiments in Section 5 demonstrate the validity of our scheme.

\subsection{Two Technologies for Training GANs}
\subsubsection{Spectral normalization for both generator and discriminator}
As we know, WGAN replaced the JS divergence with the Wasserstein divergence. However, this requires the discriminator to satisfy the Lipschitz constraint:
\begin{equation}
|f(x_1)-f(x_2)\le K|x_1-x_2|.
\end{equation}
It requires that the absolute value of the derivative of the function $f$ does not exceed $K$. Otherwise, gradient explosion will occur during model training. SNGANs satisfy the Lipschitz constraint using spectral normalization by employing the spectral normalization of each layer $\sigma (A)$ to restrict the parameter matrix $W_i$:
\begin{equation}
	\bar{W_i}=\frac{W_i}{\sigma (A)},
\end{equation}
where $\bar{W_i}$ is the parameter matrix after spectral normalization. Spectral normalization has the advantage of not requiring extra hyperparameter tuning. In addition, the computational cost is relatively less. In conclusion, it can be used to satisfy the Lipschitz constraint and stabilize the training of GANs.

In GIU-GANs, we borrowed from SAGANs: spectrum normalization is applied to both discriminator and generator to stable the training of stable GANs.

\subsubsection{WGAN-GP}
Moreover, we adopt WGAN-GP as our objective function. WGAN-GP uses a new truncation strategy, gradient penalty::
\begin{equation}
	L=E_{\tilde{x}\sim P_g}[D(\tilde{x})]-E_{\tilde{x}\sim P_r}[D(\tilde{x})]+\lambda E_{\hat{x}\sim P_{\hat{x}}} [(||\nabla_{\hat{x}}D(\hat{x})||_2-1)^2],
\end{equation}
where $P_r$ is the real image distribution, $P_g$ is the generated image distribution, $P_{\hat{x}}$ is the distribution interpolated between $P_r$ and $P_g$, and $\lambda$ is the penalty coefficient. The gradient penalty, $|\nabla_{\hat{x}}D(\hat{x})|\leq K$, satisfies the Lipschitz constraint to stabilize training of GANs. Specifically, the training details of GIU-GANs are summarized in Algorithm 2:
\begin{algorithm}[t]
	\caption{ Training Algorithm of GIU-GANs. We use default values of $\lambda$ = 10, $\alpha$ =
		0.0002, $\beta_1$ = 0, $\beta_2$ = 0.9.
} %算法的名字
	\hspace*{0.02in} {\bf Input:} The batch size m, Adam hyperparameters $\alpha$, $\beta_1$, $\beta_2$, the gradient penalty coefficient $\lambda$.\\ %算法的输入， \hspace*{0.02in}用来控制位置，同时利用 \\ 进行换行
	\hspace*{0.02in} {\bf Input:} %算法的结果输出
	Initial discriminator parameters $w_0$, initial generator parameters $\theta_0$
	\begin{algorithmic}[1]
		\For{number of training iteration} % For 语句，需要和EndFor对应
		\For{i = 1, ..., m }
		\State Sample real data $x\sim P_r$, latent variable $z\sim p_(z)$, a random number $\epsilon\sim U[0,1]$.
		\State $\tilde x\gets G_{\theta_G}(z)$.
		\State $\tilde x\gets \epsilon x+(1-\epsilon)\hat{x}$.
		\State $L^{(i)}\gets D_w(\tilde x)-D_w(x)+\lambda (||\nabla_{\hat{x}}D(\hat{x})||_2-1)^2$.
		\EndFor
		\State $w\gets Adam(\nabla_w\dfrac{1}{m}\sum_{i=1}^mL^{(i)},w, \alpha, \beta_1, \beta_2)$.
		\State Sample a batch of latent variables $\{z^{(i)}\}_{i=1}^m\sim p(z)$
		\State $\theta \gets Adam(\nabla_\theta\dfrac{1}{m}\sum_{i=1}^m-D_w(G_\theta(z)), \theta, \alpha, \beta_1, \beta_2)$
		\EndFor
		\State \Return $\theta, w$
	\end{algorithmic}
\end{algorithm}

\section{The algorithm of Representative Batch Normalization(RBN)}
The conventional BN performs three operations on input feature maps $X$: centering, scaling, and affine operations, respectively, given by
\begin{equation}
	\begin{aligned}
		&\rm{Centering}:X_m=X-E(X),\\
		&\rm{Scaling}:X_s=\frac{X_m}{\sqrt{Var(X)+\epsilon}},\\
		&\rm{Affine}:Y=X_s\gamma+\beta.
	\end{aligned}
\end{equation} 
where $E(X)$ and $Var(X)$ represent the mean and variance of feature maps, respectively, $\gamma$ and $\beta$ denote the learnable scaling and offset terms, respectively, in the affine transformation, and $\epsilon$ is used to prevent the occurrence of zero values. In the model training process, the mean and variance are calculated in a mini-batch:  
\begin{equation}
	\begin{aligned}
		&\mu_B=\frac{1}{NHW}\sum_{n=1}^N\sum_{h=1}^H\sum_{w=1}^WX_{n,c,h,w},\\
		&\sigma_B^2=\frac{1}{NHW}\sum_{n=1}^N\sum_{h=1}^H\sum_{w=1}^W(X_{n,c,h,w}-\mu_B)^2.
	\end{aligned}
\end{equation} 

However, $E(X)$ and $Var(X)$ of mini-batch use normalization to reduce the expression of partial random noise by the generator. Therefore, we replace the BN with RBN in part of the network layer of GANs. 

First, RBN introduces a centering calibration operation before the centering operation of BN:
\begin{equation}
	X_{cm(n,c,h,w)}=X_{n,c,h,w}+w_m \bigodot K_m,
\end{equation}
where $w_m\in R^{1\times C\times 1\times1}$ is the learnable variable and $K_m$ is the statistical features obtained after a series of processing of $X$. Different operations will result in different shapes of statistical features. For example, $K_m\in R^{N\times C\times 1\times1}$ is to obtain statistical information of each channel through feature map normalization, or $K_m\in R^{N\times 1\times H\times W}$ is to obtain statistical information of each pixel after channel normalization. In GIU-GANs we follow the original RBN paper \cite{DBLP:conf/cvpr/GaoHLCP21} and adopt global average pooling to normalize the feature map as the shape of $K_m\in R^{N\times C\times 1\times1}$. The new mean is $E(X_{cm})=(1+w_m)\cdot{E(X)}$, it subtracts the old mean as follows:
\begin{equation}
	\begin{aligned}
		&(X_{cm}-E(X_{cm}))-(X-E(X))\\
		&=X+w_m\cdot K_m-((1+w_m)\cdot E(X)-(X-E(X)))\\
		&=w_m\cdot (K_m-E(X)).
	\end{aligned}
\end{equation} 

After the centering calibration, $w_m $, as a learnable variable, will enhance or suppress representational noise. If $w_m \textgreater 0$ when $K_m \textgreater E(X)$, representative noise characteristics will be enhanced, whereas if $w_m \textless 0$ when $K_m \textgreater E(X)$, representative noise features will be suppressed.

Meanwhile, RBN adds a scaling calibration after the original scaling operation:
\begin{equation}
	X_{cs(n,c,h,w)}=X_{s(n,c,h,w)}\cdot R(w_v\bigodot K_s+w_b),
\end{equation}
where $w_v, w_b \in R^{1\times C\times 1\times1}$ are learnable variables, which control the strength and position of the limiting function, respectively. Because $0\textless R()\textless 1$, there must be a value $\tau$ that is $R() \textless \tau\textless 1$. Therefore, $Var(X_{cs})$ is expressed as follows:
\begin{equation}
	Var_{X_{cs}} \textless Var(X_s\tau)=\tau^2Var(X_s).
\end{equation}
After the scaling calibration, the dispersion of the characteristic variance is reduced and a more stable channel distribution is obtained.

In Section 5, we compared GIU-GANs with and without RBN on the CIFAR-10 and CelebA datasets. The experimental results revealed the effectiveness of GIU-GANs with RBN on the CelebA dataset.

\section{Experimental Results}

To evaluate the effectiveness of GIU-GANs, we performed experiments on two datasets: CelebA and CIFAR-10. The CIFAR-10 dataset comprises 50,000 and 10,000 training and testing images, respectively. We trained with all 50,000 images of the CIFAR-10 dataset. For the CelebA dataset, we aligned and cropped for 200,000 images to 64 × 64 resolution and used them as the training set. Moreover, to further verify the effectiveness of the GIU module and RBN, we added an ablation experiment in Subsection 5.4. Two common evaluation metrics, IS and FID, were used as the main metrics. IS employs pretrained InceptionNet-V3 \cite{DBLP:conf/cvpr/SzegedyVISW16} to evaluate generated images. The output of InceptionNet-V3 is a 1,000-dimensional vector. The values of each dimension of the output vector correspond to the probability that the picture belongs to a certain category. For IS, it needs to consider the quality of a single image and the diversity of a batch of images. For a single image, the entropy of the probability distribution of InceptionNet-V3 output should be minimal. Smaller means that the generated image is more likely to belong to a certain category and the image quality is high. For a batch of images generated, the average probability distribution entropy of InceptionNet-V3 output should be maximal. In other words, to ensure the diversity of images generated, GANs needs to make the output of InceptionNet-V3 spread equally across 1,000-dimensional labels. The formula for IS is expressed as follows:
\begin{equation}
	IS=exp(E_{x\in p_g}D_{KL}(P(y|x)||P(y)),
\end{equation}
where $P(y|x)$ denotes the output distribution of the generated image $x$ input to InceptionNet-V3, $P(y)$ represents the average distribution of generated images in the InceptionNet-V3 output category, and $D_KL$ means $KL$ divergence. In summary, IS is measured for a single dataset only. To measure the probability distribution of a dataset, mean and variance are crucial, so standard deviation needs to be added to determine the distribution of this dataset when calculating IS. FID measures the generated image quality by comparing the "distance" of the generated image to the real image; the lower the value the better. FID does not use the last layer of InceptioNet-V3 as the output, but the last pooled layer as the output. This layer will output a vector of 2,048 dimensions. This makes each image is predicted to have 2,048 features. Next, FID calculates the mean of the 2,048-dimensional eigenvectors and compares them with the covariance matrix. Thus, the FID score is defined as follows:
\begin{equation}
	FID(x,g)=||\mu_x-\mu_g||^2_2+Tr(\sum{_x}+\sum{_g}-2(\sum{_x}\sum{_g})^{\dfrac{1}{2}})),
\end{equation}

where $\mu_x$ and $\mu_g$ are the means of two distributions; $\sum{_x}$ and $\sum{_g}$ are the covariance of the two distributions, respectively; $Tr$ is the trace operation of the matrix, i.e., the sum of all elements of the main diagonal of the matrix. FID is used to calculate the distance between two distributions; a lower FID means a closer match between the generated image distribution and the real image distribution. It considers the mean and variance while calculating the two distributions, so there is no need to add standard deviation when calculating an FID score.

All models were trained 1 million times, and 50,000 images generated by all models were measured by PyTorch implementations of IS and FID. All experiments were set with NVIDIA RTX 3090 GPU and Intel Core i9-9900k CPU@3.60GHz using PyTorch 1.7.0. In the following, we present experimental results of GIU-GANs and compare them with other GANs.

\subsection{Hyperparameters Study}

To discuss how to insert the GIU module for the best performance of GANs, we trained different GIU-GANs on CIFAR-10; their IS and FID are summarized in Table 1. In the following table, we use "D" for discriminator, "G" for generator, "shallow" for the shallow layer of the network, and "deep" for the deep layer of the network. Figure 6 shows the insertion of different GIU modules.

\begin{figure}[H]
	\centering
	\subfigure[Middle D and G.]{
		\includegraphics[width=0.3\columnwidth]{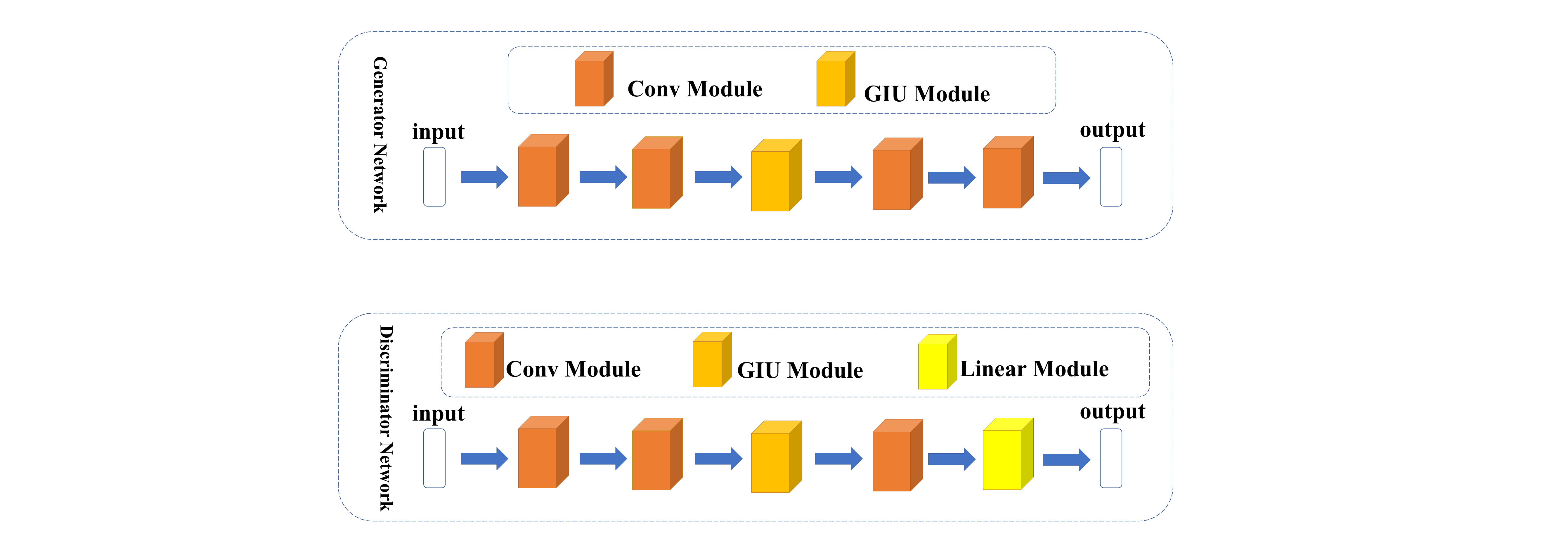}
	}
	\subfigure[Shallow D and Deep G.]{
		\includegraphics[width=0.3\columnwidth]{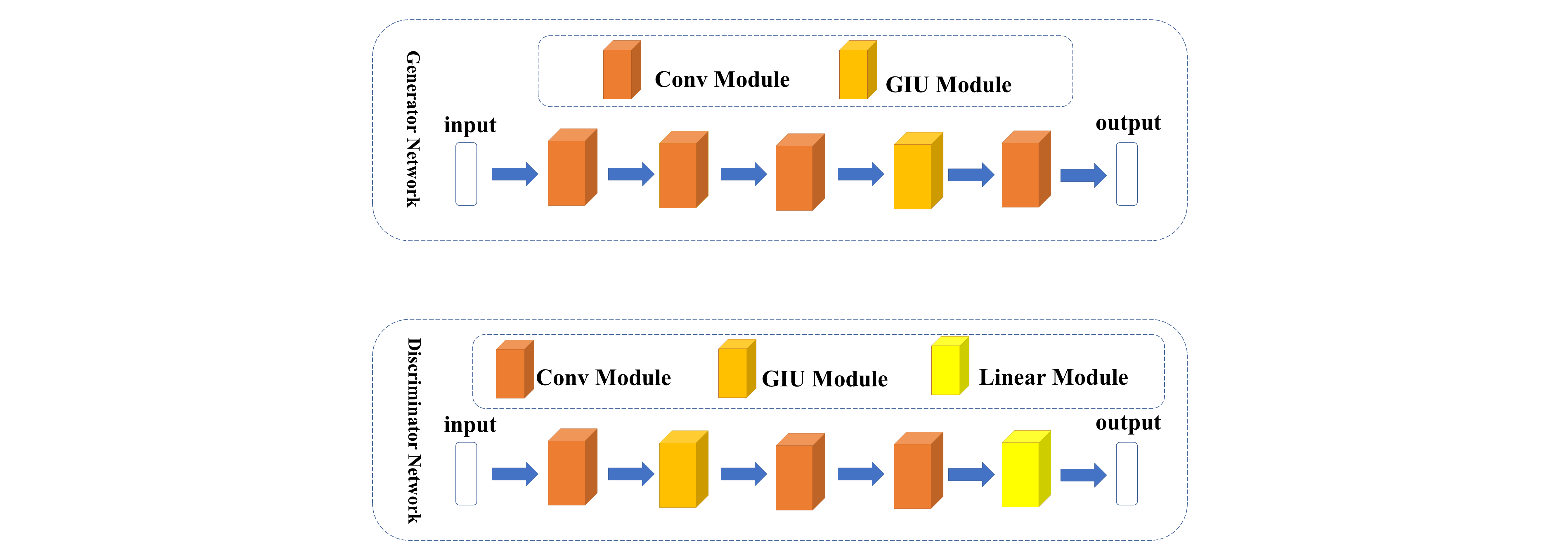}
	}
	\subfigure[Deep D and Shallow G.]{
		\includegraphics[width=0.3\columnwidth]{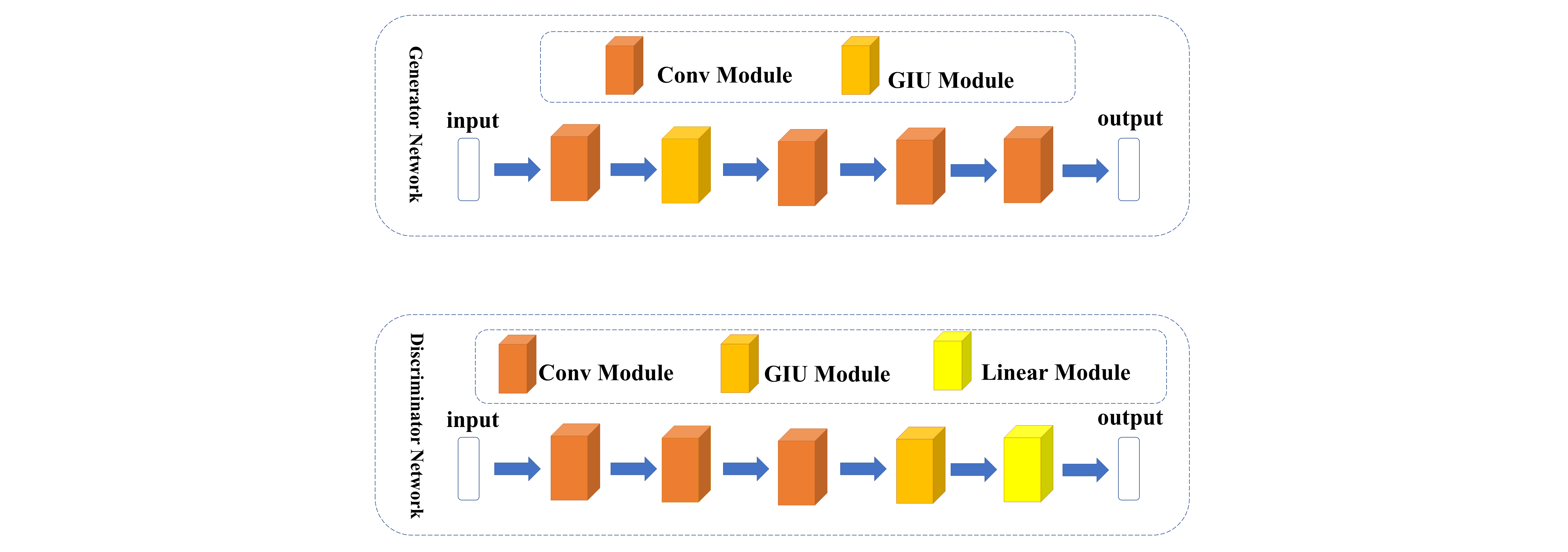}
	}
	\subfigure[All the D and G.]{
		\includegraphics[width=0.3\columnwidth]{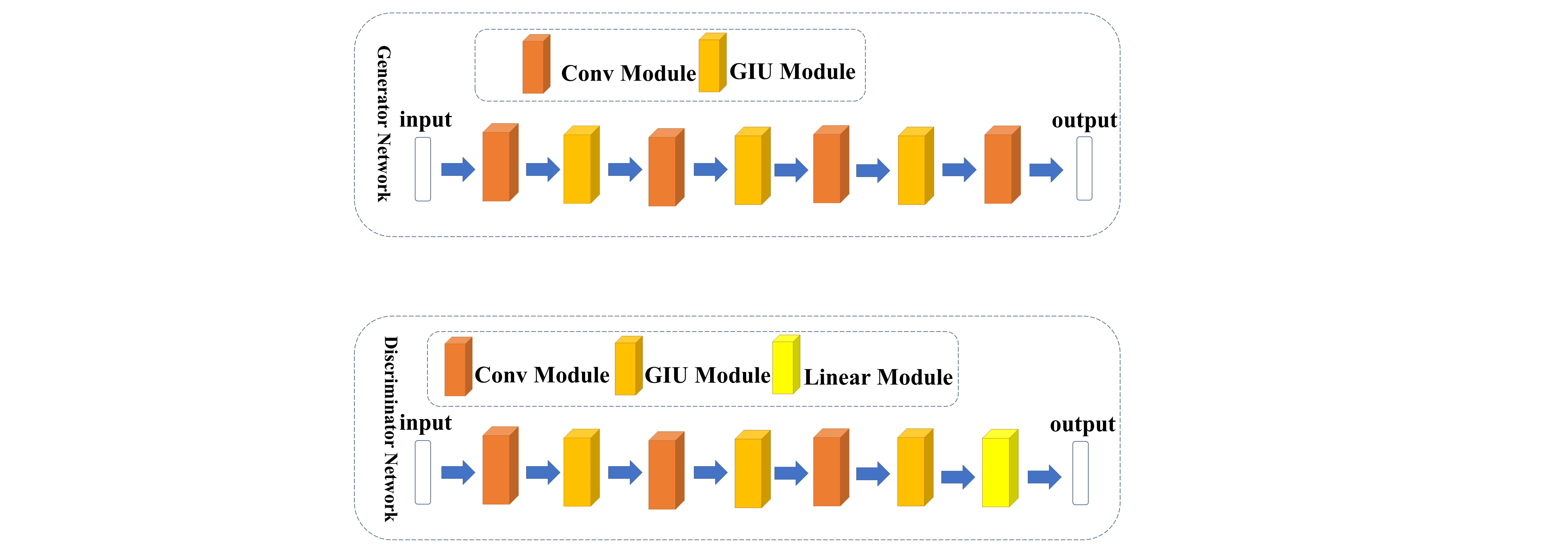}
	}
	\subfigure[Only G.]{
		\includegraphics[width=0.3\columnwidth]{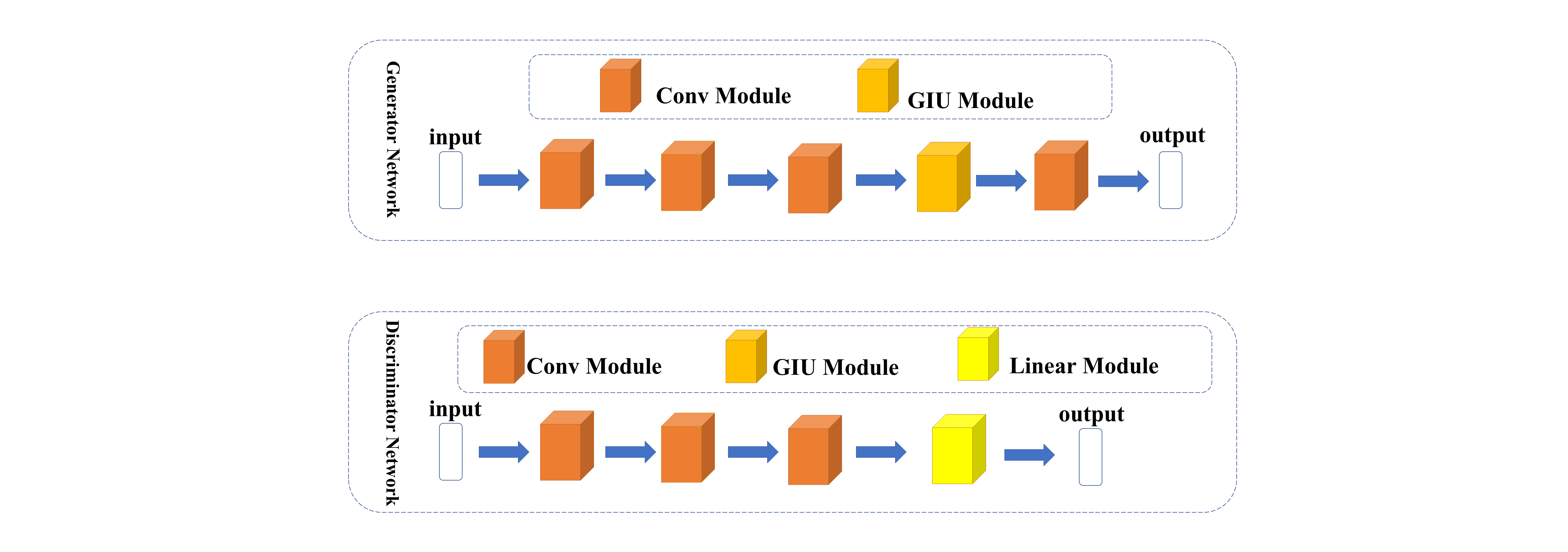}
	}
	\subfigure[Only D.]{
		\includegraphics[width=0.3\columnwidth]{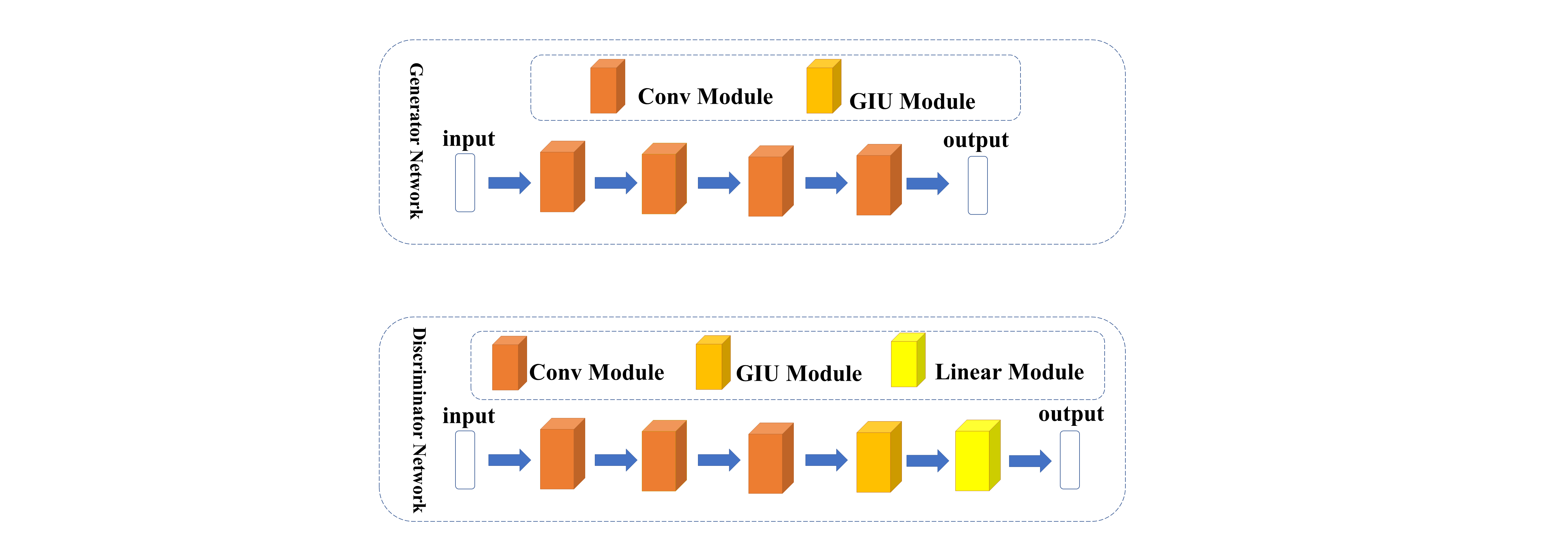}
	}

	\caption{GIU module in GANs multiple insertion situations.}
	\label{fg:clo}
\end{figure}

\begin{table}[H]
	
	\centering
	\fontsize{6.5}{8}\selectfont
	\begin{threeparttable}
		\caption{Experimental results of different locations and numbers of GIU modules on the CIFAR-10 dataset.}
		\label{tab:performance_comparison}
		\begin{tabular}{ccccccc}
			\toprule
			{Model}&FLOPs&Params&IS($\uparrow$) & FID($\downarrow$)\cr
			\midrule
			{Middle D and G}&97.24m&1.86M&$4.36 \pm 0.05$&26.72\cr
			{Shallow D and Deep G}&98.15M&1.92M&$3.96 \pm 0.04$&39.53\cr
			{Deep D and Shallow G}&97.23M&1.92M&$4.72 \pm 0.06$ & 22.66\cr
			{All the D and G}&99.03M&1.94M&$4.21 \pm 0.04$& 35.36\cr
			{Only G}&96.79M&1.88M&$4.52 \pm 0.06$&26.24\cr
			{Only D}&96.79M&1.88M&$4.44 \pm 0.07$&29.84\cr
			\bottomrule
		\end{tabular}
	\end{threeparttable}
\end{table}

Then, to verify which hyperparameters are suitable for our model, different hyperparameters were applied to GIU-GANs and trained on the CIFAR-10 dataset. We list the details of these hyperparameters in Table 2 and record the experimental results in Table 3.

\begin{table}[H]
	
	\centering
	\fontsize{6.5}{8}\selectfont
	\begin{threeparttable}
		\caption{Table of different GIU-GANs parameter settings.}
		\label{tab:performance_comparison}
		\begin{tabular}{ccccccc}
			\toprule
			{Model}&$D_{lr}$&$G_{lr}$&Batch Size&Optimization Algorithm&FLOPs&Params\cr
			\midrule
			GIU-GANs-V1&0.0001&0.0004&64&Adam\cite{DBLP:journals/corr/KingmaB14}&97.23M&1.92M\cr
			GIU-GANs-V2&0.0003&0.0003&64&Adam&97.23M&1.92M\cr
			GIU-GANs-V3&0.0002&0.0002&64&Adam&97.23M&1.92M\cr
			GIU-GANs-V4&0.0002&0.0002&64&Adagrad \cite{DBLP:journals/jmlr/DuchiHS11}&97.23M&1.92M\cr
			GIU-GANs-V5&0.0002&0.0002&64&RMSprop&97.23M&1.92M\cr
			\bottomrule
		\end{tabular}
	\end{threeparttable}
\end{table}

\begin{table}[H]
	
	\centering
	\fontsize{6.5}{8}\selectfont
	\begin{threeparttable}
		\caption{Experimental results of GIU-GANs with different parameter settings on CIFAR-10.}
		\label{tab:performance_comparison}
		\begin{tabular}{ccccccc}
			\toprule
			{Model}&FLOPs&Params&IS($\uparrow$) & FID($\downarrow$)\cr
			\midrule
			{GIU-GANs-V1}&97.23M&1.92M&$4.39 \pm 0.12$ & 34.80\cr
			{GIU-GANs-V2}&97.23M&1.92M&$4.78 \pm 0.06$ & 23.18\cr
			{GIU-GANs-V3}&97.23M&1.92M&$4.72 \pm 0.06$ & 22.66\cr
			{GIU-GANs-V4}&97.23M&1.92M&$3.25 \pm 0.03$ & 61.28\cr
			{GIU-GANs-V5}&97.23M&1.92M&$4.47 \pm 0.06$ & 28.93\cr
			\bottomrule
		\end{tabular}
	\end{threeparttable}
\end{table}

As can be seen from the IS and FID of the model participating in the experiment, GIU-GANs-V3 should be selected for setting the hyperparameters of GIU-GANs.

Meanwhile, different $K$ implies that the involution kernel has different sizes, and different $G$ represents different numbers of channels sharing the kernel. In our experiments, we use the hyperparameter $Group\_Channels$s to set the $G=\dfrac{C_i}{Group\_Channels}$. From the parameter formula of the GIU module $\dfrac{(C_i^2+K^2 GC_i)}{r_1}+2C_i+\dfrac{(2C_i^2)}{r_2}$, we know that different $K$ and $G$ indicate different numbers of parameters; to further explore their effects on GIU-GANs, we set up GIU-GANs with different $K$ and $Group\_Channels$ on the CIFAR-10 dataset for an experiment. The results are summarized in Table 4.

\begin{table}[H]
	
	\centering
	\fontsize{6.5}{8}\selectfont
	\begin{threeparttable}
		\caption{Effects of different $K$ and $Group\_Channels$ for GIU-GANs on CIFAR-10.}
		\label{tab:performance_comparison}
		\begin{tabular}{ccccccc}
			\toprule
			{$K$}&$Group\_Channels$&Params&FLOPs&IS($\uparrow$) & FID($\downarrow$)\cr
			\midrule
			{3}&16&97.23M&1.92M&$4.72 \pm 0.06$ & 22.66\cr
			{5}&16&97.76M&1.95M&$4.64 \pm 0.06$ & 24.71\cr
			{7}&16&98.56M&1.99M&$4.73 \pm 0.03$ & 22.97\cr
			{3}&8&97.52M&1.93M&$4.58 \pm 0.06$ & 24.50\cr
			{3}&32&97.08M&1.91M&$4.65 \pm 0.06$ & 24.89\cr
			\bottomrule
		\end{tabular}
	\end{threeparttable}
\end{table}

Through experimental comparison, Adam was set as the optimization algorithm of GIU-GANs; the generator and discriminator’s learning rates were set to 0.0002; $K$ should be set to 3, and $Group\_Channels$ should be 16.

\subsection{Results on the CIFAR-10 Dataset}

The architecture of the GIU-GANs used to train the CIFAR-10 dataset is shown in Figure 7:

\begin{figure}[H]
	\centering
	\subfigure[Discriminator Network.]{
		\includegraphics[width=0.45\columnwidth]{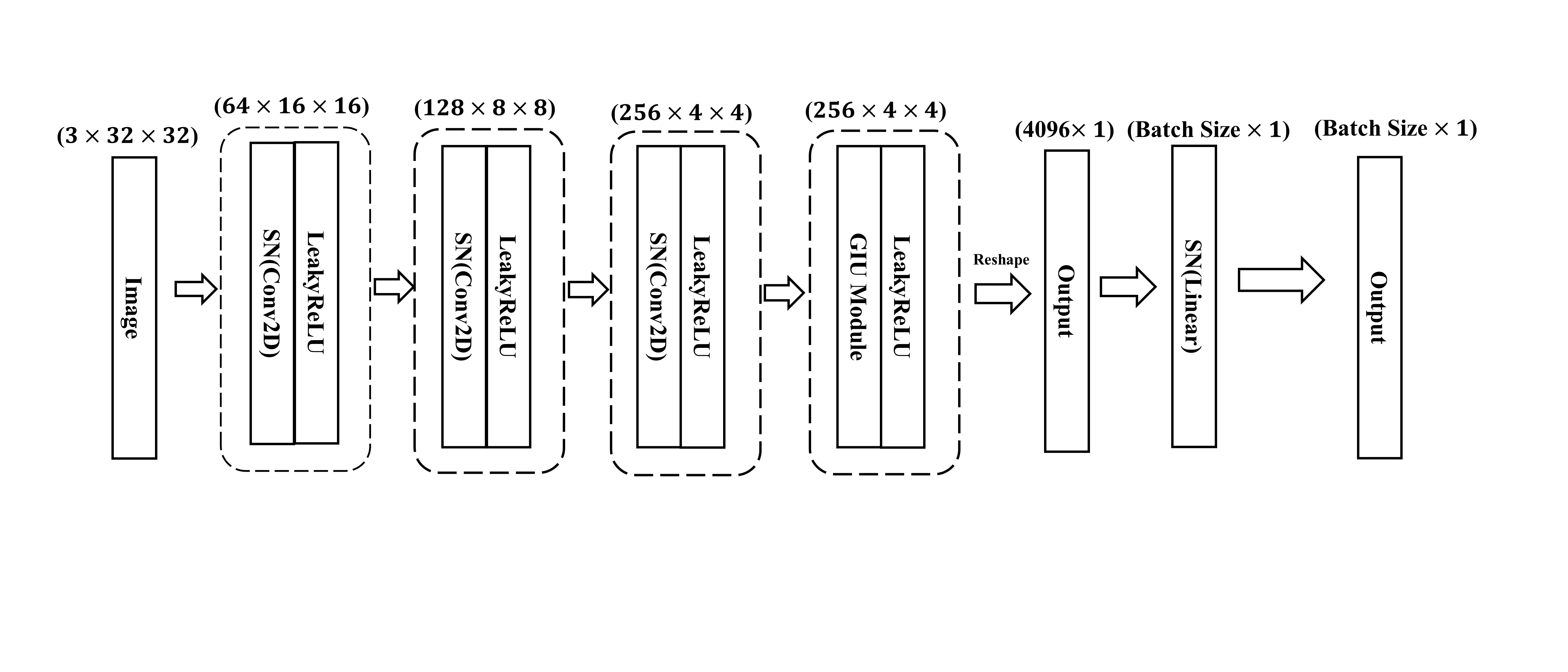}
	}
	\subfigure[Generator Network.
	]{
		\includegraphics[width=0.45\columnwidth]{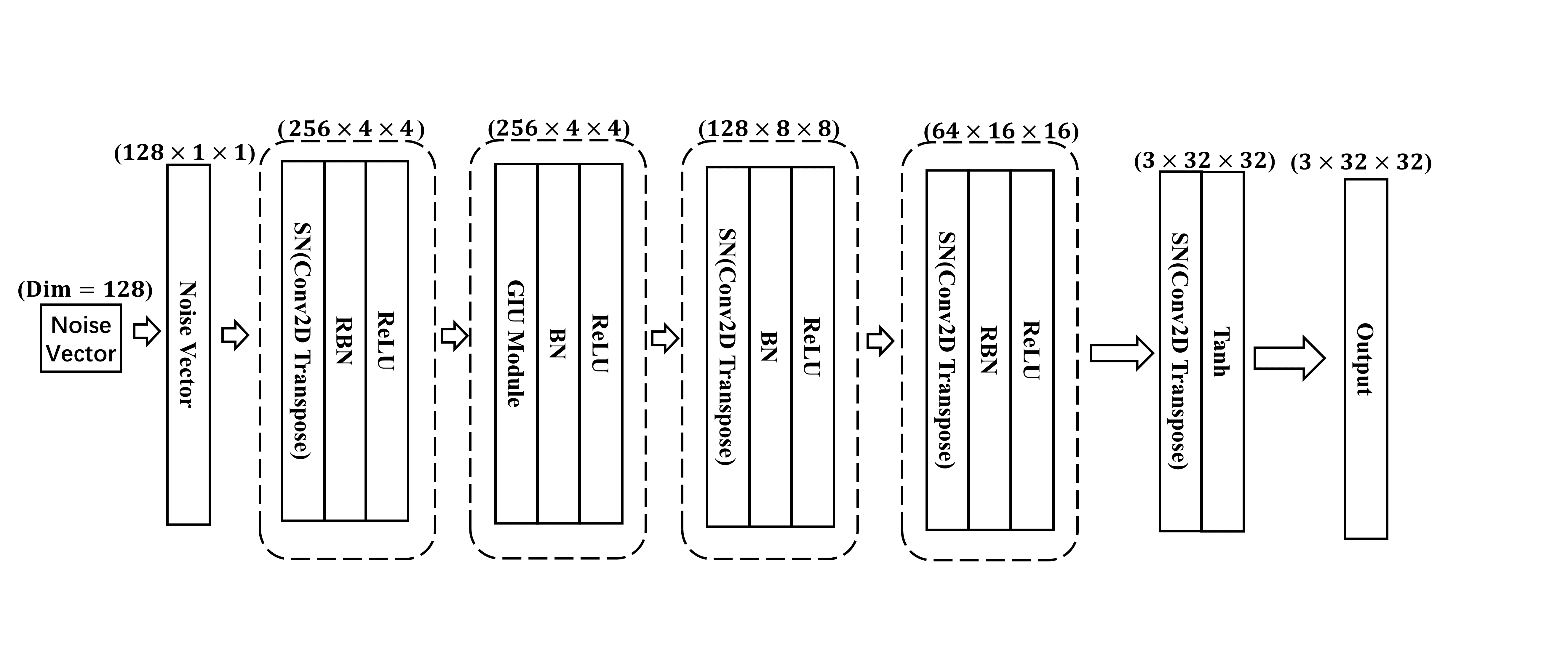}
	}
	\caption{The architecture of GIU-GANs applied on CIFAR-10.}
	\label{fg:clo}
\end{figure}

In this experiment, GIU-GANs and other counterpart GANs were all trained 1 million times on the CIFAR-10 dataset. All GANs generated 32 × 32 images, and we calculated their IS and FID scores. The best results of all models are shown in Table 5.

\begin{table}[H]
	
	\centering
	\fontsize{6.5}{8}\selectfont
	\begin{threeparttable}
		\caption{IS and FID scores of DCGANs, WGAN-GP, LSGANs, SNGANs, SNGANs and GIU-GANs on CIFAR-10.}
		\label{tab:performance_comparison}
		\begin{tabular}{ccccccc}
			\toprule
			{Model}&FLOPs&Params&IS($\uparrow$) & FID($\downarrow$)\cr
			\midrule
			{DCGANs}&309.24M&4.34M&$4.34 \pm 0.05$ &26.22\cr
			{WGAN-GP}&96.39M&1.85M&$4.51 \pm 0.06$ &22.95\cr
			{LSGANs}&117.25M&1.15M&$3.24 \pm 0.03$ &66.51\cr
			{SNGANs}&96.36M&1.85M&$4.40 \pm 0.04$ &24.60\cr
			{SAGANs}&101.65M&1.98M&$3.52 \pm 0.03$ &67.31\cr
			\textbf{GIU-GANs}&\textbf{97.23M}&\textbf{1.92M}&\bm{$4.72 \pm 0.06$} &\textbf{22.66}\cr
			\bottomrule
		\end{tabular}
	\end{threeparttable}
\end{table}

Figure 8 shows 64 images generated by GIU-GANs; from Table 5, GIU-GANs outperformed other well-known GANs.

\begin{figure}[H]
	\centering
	\includegraphics[width=5cm,height=5cm]{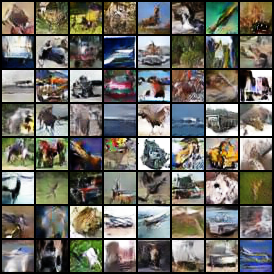}
	\vspace{-1.0em}
	\caption{32×32 examples randomly produced by GAN models on CIFAR-10.}
\end{figure}

\subsection{Results on the CelebA Dataset}

The architecture of the GIU-GANs used to train the CelebA dataset is shown in Figure 9:
\begin{figure}[H]
	\centering
	\subfigure[Discriminator Network.]{
		\includegraphics[width=0.45\columnwidth]{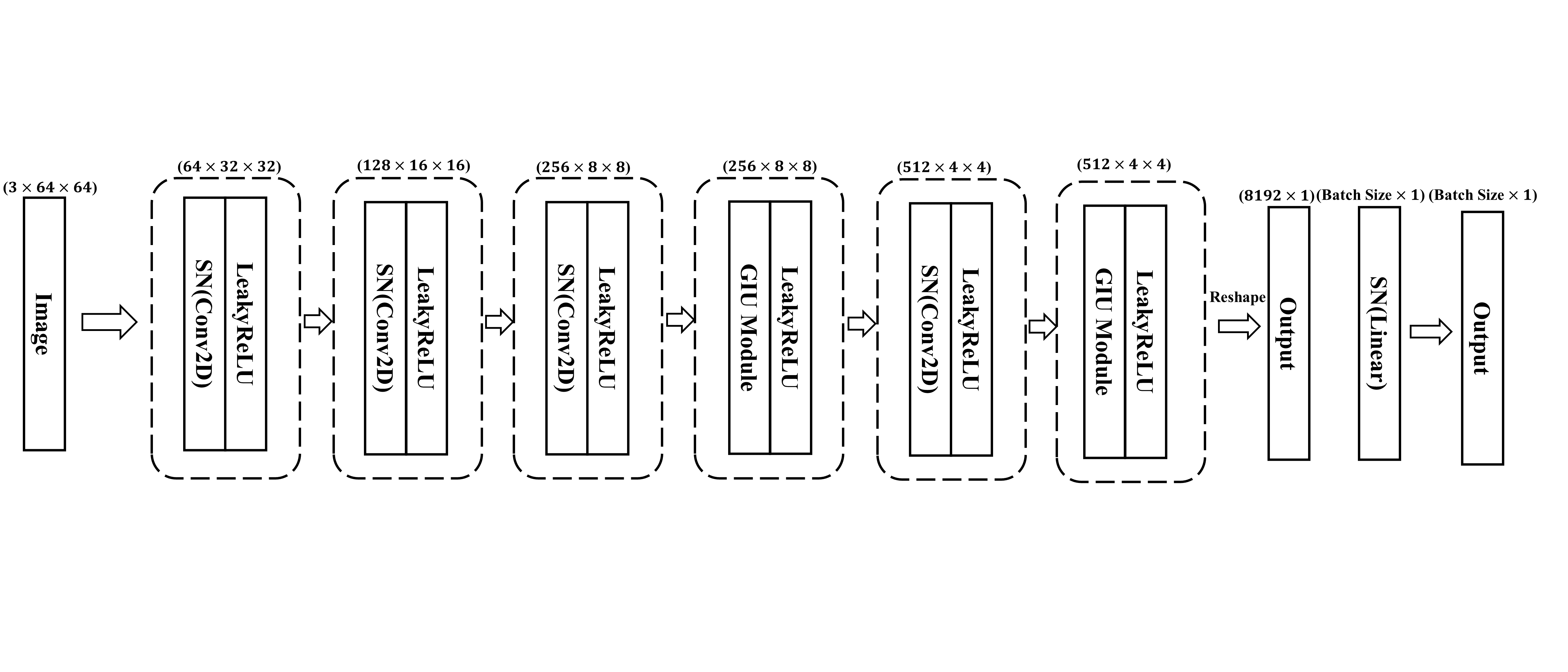}
	}
	\subfigure[Generator Network.
	]{
		\includegraphics[width=0.45\columnwidth]{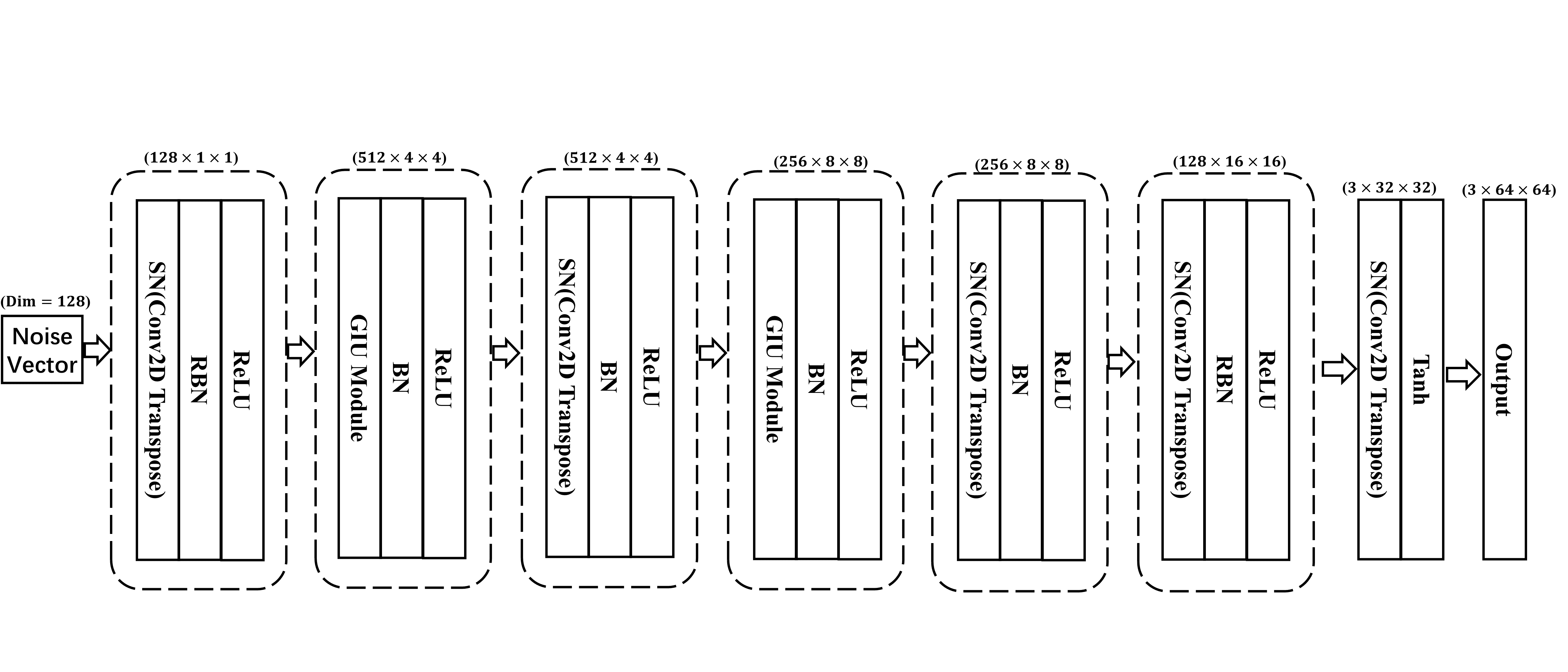}
	}
	\caption{The architecture of GIU-GANs applied on CelebA.}
	\label{fg:clo}
\end{figure}

In this experiment, we cropped and aligned all images from the CelebA dataset for training and trained 1 million times. All GANs generated 50k 64×64 images, and we calculated their IS and FID scores.The best results of all models are shown in Table 6.

In addition, as shown in Figure 10, when DCGANs were trained on the CelebA dataset, the gradient-vanishing of the discriminator resulted in no change of generator gradient. Thus, we no longer used DCGANs for comparison

\begin{figure}[H]
	\centering
	\subfigure[Convergence curve of the discriminator of DCGANs on CelebA.]{
		\includegraphics[width=0.45\columnwidth]{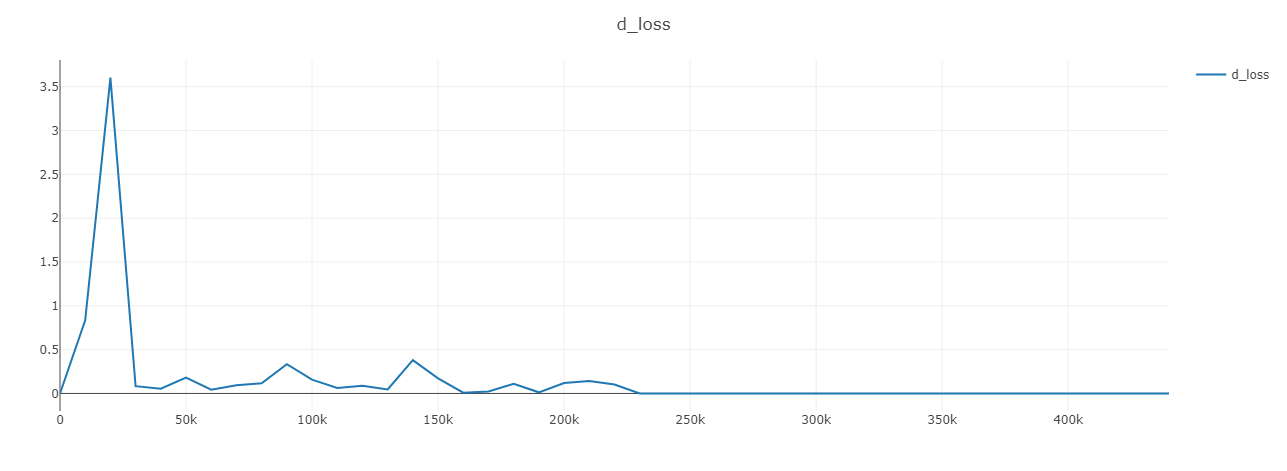}
	}
	\subfigure[Convergence curve of the generator of DCGANs on CelebA.]{
		\includegraphics[width=0.45\columnwidth]{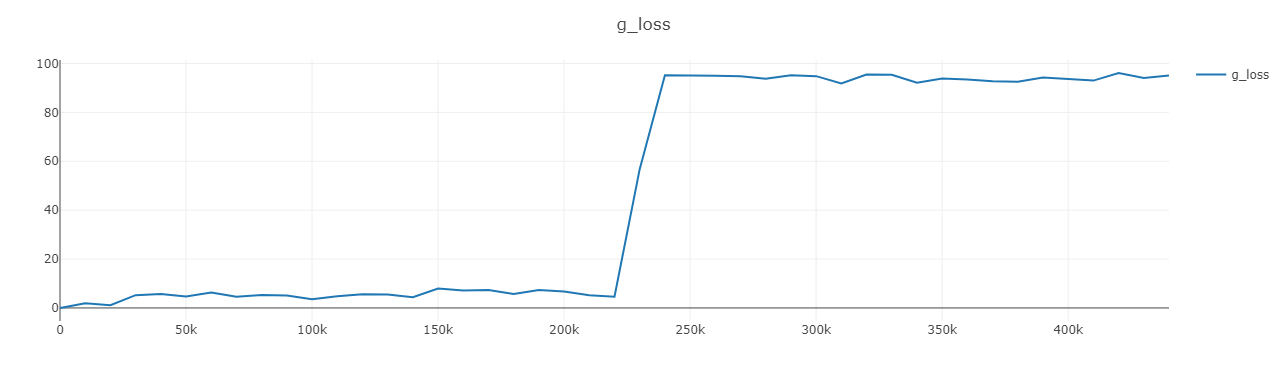}
	}
	\caption{Convergence curve of DCGANs on CelebA.}
	\label{fg:clo}
\end{figure}

\begin{table}[H]
	
	\centering
	\fontsize{6.5}{8}\selectfont
	\begin{threeparttable}
		\caption{IS and FID scores of WGAN-GP, LSGANs, SNGANs, SNGANs and GIU-GANs on CelebA.}
		\label{tab:performance_comparison}
		\begin{tabular}{ccccccc}
			\toprule
			{Model}&FLOPs&Params&IS($\uparrow$) & FID($\downarrow$)\cr
			\midrule
			{WGAN-GP}&536.58M&6.57M&$2.48 \pm 0.03$ &11.07\cr
			{WGAN-GP(+2)}&687.72M&12.47M&$2.50 \pm 0.02$ &12.39\cr
			{WGAN-GP(+3)}&764.48M&12.66M&$2.57 \pm 0.02$ &8.98\cr
			{LSGANs}&694.48M&1.89M&$2.40 \pm 0.02$ &15.48\cr
			{SNGANs}&536.48M&6.57M&$2.54 \pm 0.02$ &7.09\cr
			{SAGANs}&557.58M&7.00M&$2.11 \pm 0.02$ &19.10\cr
			\textbf{GIU-GANs}&\textbf{543.28M}&\textbf{6.91M}&\bm{$2.61 \pm 0.02$} &\textbf{6.34}\cr
			\bottomrule
		\end{tabular}
	\end{threeparttable}
\end{table}

The images generated by GIU-GANs are shown in Figure 11. From Table 6, GIU-GANs performed best in both IS and FID.

\begin{figure}[H]
	\centering
	\includegraphics[width=5cm,height=5cm]{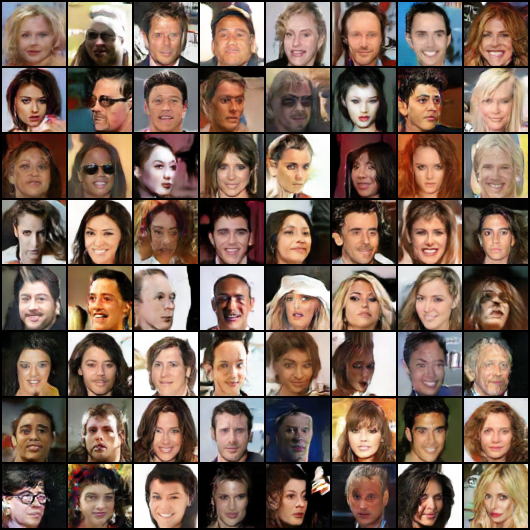}
	\vspace{-1.0em}
	\caption{64×64 examples randomly produced by GAN models on CelebA.}
\end{figure}

\subsection{Ablation Study}

To explore the reasons for the superior performance of GIU-GANs, several ablation studies were designed to study the contributions of individual components. We applied the GIU-GANs with and without the GIU module on the CIFAR-10 dataset. In addition, we employed WGAN-GP with spectral normalization for the generator and discriminator as the baseline. The best results are shown in Table 7.

\begin{table}[H]
	
	\centering
	\fontsize{6.5}{8}\selectfont
	\begin{threeparttable}
		\caption{Experimental results of GIU module ablation.}
		\label{tab:performance_comparison}
		\begin{tabular}{ccccccc}
			\toprule
			{Model}&FLOPs&Params&IS($\uparrow$) & FID($\downarrow$)\cr
			\midrule
			{baseline}&96.36M&1.85M&$3.83 \pm 0.05$ &37.54\cr
			{GIU-GANs without GIU module}&96.36M&1.85M&$3.99 \pm 0.05$ &39.97\cr
			\textbf{GIU-GANs}&97.23M&1.92M&\bm{$4.72 \pm 0.02$} &\textbf{22.66}\cr
			\bottomrule
		\end{tabular}
	\end{threeparttable}
\end{table}

From Table 7, the performance of GIU-GANs deteriorated significantly when the GIU module was removed.

Table 8 shows the influence of RBN on GIU-GANs. GIU-GANs without RBN on the CIFAR-10 dataset performed well; however, for the CelebA dataset, GIU-GANs with RBN was superior.

\begin{table}[H]
	\caption{Experimental results of RBN ablation.}
	\resizebox{\textwidth}{15mm}{
		
		\begin{tabular}{ccccccccc}
			\toprule
			\multirow{2}{*}{Method}&
			\multicolumn{4}{c}{CIFAR-10}&\multicolumn{4}{c}{CelebA}\cr
			\cmidrule(lr){2-5} \cmidrule(lr){6-9}
			&FLOPs&Params&IS($\uparrow$) & FID($\downarrow$)&FLOPs&Params&IS($\uparrow$) & FID($\downarrow$)\cr
			\midrule
			GIU-GANs without RBN&97.22M&1.92M&$4.79 \pm 0.03$ & 22.12&543.28M&6.91M&$2.50 \pm 0.03$ & 6.53\cr
			\textbf{GIU-GANs}&97.23M&1.92M&\bm{$4.72 \pm 0.06}$ &\textbf{22.66}&543.28M&6.91M&\bm{$2.61 \pm 0.02$} &\textbf{6.34}\cr
			\bottomrule
		\end{tabular}}
\end{table}

\subsection{Results Analysis}

The training of GANs is extremely unstable. After a long time of training, the error increases, resulting in the overfitting phenomenon. In GANs, some hidden layers have thousands of channels, and redundant channel information not only increases the computational time of the model but also makes the model more complex and increases the overfitting risk. The GIU module is placed after the hidden layer with a large number of channels. The GIU module first filters the global information, enhances useful information, and suppresses useless information. Then, the GIU module pays attention to each pixel of the feature maps and generates the corresponding parameter matrix for multiplication--addition. By filtering the global channels and paying attention to each pixel on the feature maps, the GIU module can exploit global information. The performance of GIU-GANs can be improved without stacking convolutions, so GIU-GANs can easily enhance the generated image quality without much consideration of the overfitting risk.

Moreover, WGAN-GP achieved results similar to GIU-GANs on the CIFAR-10 dataset. We argued that the computational complexity of WGAN-GP is FLoating-point OPerations (FLOPs): 96.39M and parameters: 1.85M. The computational complexity of our model is FLOPs: 97.23M and parameters: 1.92M. Models with lower complexity are less prone to overfitting when handling low fractional rate images, such as CIFAR-10 (32×32), compared with models with higher complexity. Meanwhile, in processing higher resolution images, such as CelebA (64×64), the less complex WGAN-GP (IS: $2.48 \pm 0.03$ and FID:11.07) does not perform as well as the more complex GIU-GANs (IS: $2.61 \pm 0.02$ and FID:6.34)

\section{Conclusion}

In this study, we proposed a new generative model called GIU-GANs, which incorporates a GIU module into the GANs framework. GIU-GANs place the GIU module comprising SE module and involution so that the model can exploit channel information to extract features and reduce redundant information to enhance the generated image quality. In addition, because RBN pays attention to the representative feature, we replace the BN of the generator near the input and output layers with RBN. Comparing GIU-GANs with other GANs, the IS and FID of GIU-GANs reached the state-of-the-art level on CIFAR-10 and CelebA datasets.

Considering the success of the Vision Transformer (VIT) \cite{DBLP:conf/iclr/DosovitskiyB0WZ21} model in visual tasks, we plan to combine the VIT model with GANs in the future to further improve performance. Moreover, involution does not change the number of output channels for the GIU-GANs to abandon convolution. How to improve involution so that GANs can neglect convolution and build a backbone only with involution is challenging and is considered future work.

\section{Acknowledgements}
This research work is supported in part by the Fundamental Research Funds for the Central Universities under Grant 21621017, in part by the Innovative Youth Program of Guangdong University under Grant 2019KQNCX194, and in part by the Educational and Scientifical Project of Guangdong Province 2021GXJK368.

\bibliography{mybibfile}

\end{document}